\documentclass{llncs}
\usepackage{graphicx}
\usepackage{tabularx}
\usepackage{adjustbox}

\usepackage[table,xcdraw]{xcolor}

\begin{document}

\title{Estimating Achilles tendon healing progress\\with convolutional neural networks}

\author{Norbert Kapinski\inst{1}, Jakub Zielinski\inst{1,3}, Bartosz A. Borucki\inst{1}, Tomasz Trzcinski\inst{2,5}, Beata Ciszkowska-Lyson\inst{4}, Krzysztof S. Nowinski\inst{1}}

\institute{$^1$University of Warsaw $^2$Warsaw University of Technology\\
$^3$Medical University of Warsaw $^4$Carolina Medical Center $^5$Tooploox}

\maketitle              

\begin{abstract}
Quantitative assessment of a treatment progress in the Achilles tendon healing process - one of the most common musculoskeletal disorders in modern medical practice - is typically a long and complex process: multiple MRI protocols need to be acquired and analysed by radiology experts for proper assessment. In this paper, we propose to significantly reduce the complexity of this process by using a novel method based on a pre-trained convolutional neural network. We first train our neural network on over 500 000 2D axial cross-sections from over 3 000 3D MRI studies to classify MRI images as belonging to a healthy or injured class, depending on the patient’s condition. We then take the outputs of a modified pre-trained network and apply linear regression on the PCA-reduced space of the features to assess treatment progress. Our method allows to reduce up to 5-fold the amount of data needed to be registered during the MRI scan without any information loss. Furthermore, we are able to predict the healing process phase with equal accuracy to human experts in 3 out of 6 main criteria. Finally, contrary to the current approaches to healing assessment that rely on radiologist subjective opinion, our method allows to objectively compare different treatments methods which can lead to faster patient’s recovery.

\end{abstract}
\section{Introduction}
Injuries of the Achilles tendons are one of the most common musculoskeletal disorders in modern medical practice, with more than 18 case per 100,000 people per year~\cite{Raikin14}. A ruptured tendon undergoes a surgical reconstruction followed by a rehabilitation process. A risk of the tendon re-rupture is between 20-40\% and proper oversight of the healing process is needed to increase the chances that the tendon is not ruptured again. Furthermore, continuous monitoring of the healing process can be useful for perfecting treatment techniques and adjusting them to patient's personal conditions.	 

Recently, several works showed how  quantitative methods based on deep neural networks can successfully be used to monitor the healing of the Achilles tendon~\cite{Kapinski2017,Nowosielski17}. However, those methods are dedicated to simple classification tasks and do not fully take advantage of the mid-level neural network representation. As suggested by~\cite{Oquab14}, those representations can be useful, especially for medical image processing tasks~\cite{Doe17,Bar2015}.

In this paper, we introduce a novel method for continuous evaluation of reconstructed Achilles tendon healing based on the responses of intermediate convolutional neural network layers. First, we train a neural network to classify MRI data as ‘healthy’ or ‘injured’. We then use the pre-trained network as a feature extractor and build the representation of the image data from the outputs of the first fully connected layer. After reducing the dimensionality of this representation with Principal Component Analysis (PCA), we use linear regression to fit the resulting representation to the scores assigned by human annotators that describe a healing phase. The obtained results indicate that the proposed approach allows to reduce the number of MRI protocols used for healing assessment from 10 to 2, without losing accuracy of the healing phase classification. Furthermore, using the responses of the proposed neural network, we are able to estimate the healing phase measured on a five-point scale with a mean square error of less then a half point, when compared with human experts.


\section{Method}

In this section, we present our method that relies on the outputs of the intermediate neural network layers to predict the healing process phase. We start by training the AlexNet architecture~\cite{AlexNet} for the binary classification task of assigning ‘healthy’ {\it vs} ‘injured’ label to the input MRI images. Once trained, we use the truncated version of this architecture as a feature extractor. More precisely, we use the outputs of the first fully connected layer (fc6) and apply a Principal Component Analysis to reduce the dimensionality of the resulting representation. Finally, we introduce a metric $H$ which is a novel contribution of the paper, representing a score of the Achilles tendon condition, visible in a single 3D MRI study image of a given protocol:
\begin{equation}
H = TM(PC1(x_1), PC1(x_2),..., PC1(x_n))
\end{equation}
where TM is a truncated mean with 2.5 upper and lower hinges (a value we obtained in our initial set of experiments), $PC1(x_i)$ is the first principal component value for the network inference performed on the slice $x_i$ where $i$ is the index of the slice in the 3D MRI study. 

\begin{figure}[t!]
\centering
\includegraphics[width=0.9\textwidth]{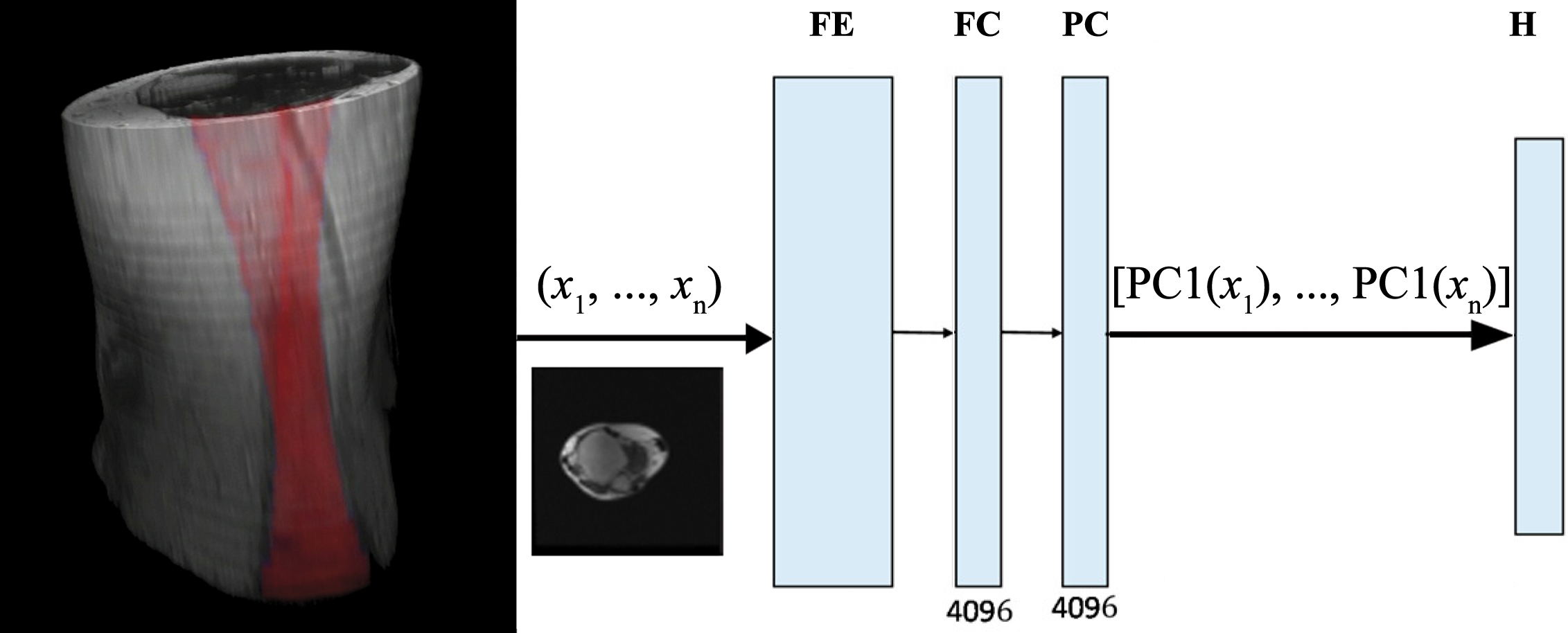}
\caption{The neural network topology used to generate tendon healing assessment.}\label{fig:net}
\end{figure}

Fig.~\ref{fig:net} shows the overview of our framework based on a neural network.
We take the feature extractor (FE) and fully connected layer (FC) from the pre-trained AlexNet model. The network uses features extracted for healthy tendons and those after rupture in different time of the healing. The FC layer performs initial weighting of the features, yet avoiding strong discretization that increases with the subsequent fully connected layers of the binary trained AlexNet model. The PC layer, comprising of Principal Components, reduces the number of output parameters while preserving the differentiation of the tendons in different healing state. Finally, $H$ reduces the number of outliers, as presented in details in Sec. 3.2, with the use of truncation and produces a single value output for the whole 3D MRI study of a given protocol.

\section{Experiments}

In this section, we present experimental results obtained with the proposed method. First, we describe the dataset used in our experiments and show the results obtained using classification algorithms based on several neural network architectures. We then verify if the $H$ metric proposed in this work corresponds to the progress of the healing process. Finally, we evaluate our approach and present studies on minimal number of MRI protocols required for the healing process assessment. 

\subsection{Dataset}

We acquired MRI data of a lower limb of healthy volunteers and patients after the Achilles tendon rupture with the use of a GE Signa HDxt 1.5T scanner with Foot \& Ankle array coil. 

To monitor the progress of the healing, each of the individuals was scanned with 10 MRI protocols, typically used in orthopaedics (and containing the most significant visual information) i.e. four 3D FSPGR Ideal [Fast Spoiled Gradient Echo] (In Phase, Out Phase, Fat, Water), PD [Proton Density], T1, T2, T2 mapping, T2$^\ast$ GRE [Gradient Echo] and T2$^\ast$ GRE TE\_MIN [Minimal Time Echo]. 
The group of healthy patients was scanned only once, while the injured patients were scanned once before the tendon reconstruction surgery and 9 times afterwards (after 1, 3, 6, 9, 12, 20, 26, 40 and 52 weeks). As of March 2018, we collected 270 3D MRI scans of healthy individuals (including 27 volunteers, 10 protocols) and 2 772 of injured patients (including 60 patients, 10 protocols and up to 10 timesteps). 

To overcome the possible artifacts of volumetric interpolation with spatially anisotropic data resolution and to represent the internal structure of tendon tissue, we performed an additional analysis only within axial slices of the 3D MRI scans. We augmented the healthy tendon slices number by mirroring and random rotations in the range of -10 to 10 degrees. We also mirrored the slices representing injured tendons. The resulting dataset contains 234,502 slices labeled as healthy and 277,208 slices labeled as injured.

To determine the ground truth for our dataset we worked with a group of expert radiologists. More precisely, we prepared a survey that includes 6 parameters describing the tendon healing process, visible in the MRI scans: 
 
\begin{enumerate}
\item Structural changes within the tendon (SCT) - informs about the loss of cohesion within the tendon area.
\item Tendon thickening (TT) - informs about the maximum dimension in the sagittal direction.
\item Sharpness of the tendon edges (STE) - informs about the edge fractality.
\item Tendon edema (TE) - informs about an abnormal accumulation of fluid in the interstitium of the tendon.
\item Tendon uniformity (TU) - informs about the level of similarity of subsequent cross-sections of the tendon.
\item Tissue edema (TisE) - informs about an abnormal accumulation of fluid and enlarged size of the fascial compartment.
\end{enumerate}

We asked experts to assess the healing condition of Achilles tendons based on the MRI images from our dataset. For each of the parameters, the experts could select a single score in a 5 point scale, where 1 describes a healthy tendon and 5 describes severely injured one. 
\subsection{Tendon healing process assessment}
%
\subsubsection{Binary classification task:}

We first evaluate neural networks on the task of binary classification of the dataset. Tab. 1 presents the comparison of 5-fold cross-validation accuracy results of AlexNet, ResNet-18 \cite{ResNet} and GoogleNet \cite{GoogleNet} models used to classify the tendon condition (healthy vs. injured). 3 folds were used for training, one for validation and one for testing.
\begin{table}[t!]
\setlength{\tabcolsep}{14pt}
\centering
\caption{5-fold cross-validation accuracy results for the tendon binary classification.}
\label{cross-validation}
\begin{tabular}{
>{\columncolor[HTML]{FFFFFF}}l |
>{\columncolor[HTML]{FFFFFF}}l |
>{\columncolor[HTML]{FFFFFF}}l |
>{\columncolor[HTML]{FFFFFF}}l |
>{\columncolor[HTML]{FFFFFF}}l }
{\color[HTML]{000000} }          & Average   & Min   & Max   & SD   \\ \hline \hline
{\color[HTML]{000000} AlexNet}   & 99.19 & 99.15 & 99.24 & 0.04 \\ \hline
{\color[HTML]{000000} ResNet-18} & 95.98 & 92.78 & 99.04 & 2.5  \\ \hline
{\color[HTML]{000000} GoogleNet} & 99.83 & 99.68 & 99.91 & 0.1  \\ 
\end{tabular}
\end{table}

The performance of all of the tested models on the classification task is satisfactory. Nevertheless, the average, minimum and maximum accuracy for both AlexNet and GoogleNet reaches over 99\% and in case of ResNet-18 the results are scattered between 93 and 99\%, giving an average accuracy of 96\%.

The AlexNet is the least complex of the tested models and its training takes less than two hours, while in the case of ResNet-18 it takes over 20 hours to train the network, for GoogleNet this time increases to around 48h. All times were measured on a server station with NVIDIA V100 GPU. Taking into account the accuracy and the model complexity, we choose the AlexNet architecture for all the experiments in the remainder of this paper.

Most of the samples labeled as injured and misclassified by our network as healthy come from 3D MRI scans that focus on the area far away from the rupture. In this work, we eliminate those outliers by using a truncated mean of our $H$ metric, but in future work we plan to address this problem by incorporating fuzzy logic or soft classification methods.

\subsubsection{Principal Components Analysis:}
For this task, we use the feature extractor part of the binary trained AlexNet model and its first fully connected layer (see Fig. 1). We evaluate whether the number of outputs can be decreased while preserving the differentiation of the stages of the healing process. For this purpose we use 48,225 slices derived from 10 patients that concluded a full year of the rehabilitation and were monitored in 10 timesteps. We perform PCA and compare the amount of variance preserved by 1, 10 and 200 most significant principal components. 
First 1, 10 and 200 principal components preserve 50.2, 90.8 and 98.8\% amount of variance respectively. This means that 4096 outputs of our modified AlexNet topology can be reduced to a single output while preserving over 50\% of variance. We therefore decide to test our approach using only the most significant principal component PC1 and this approach is used in the remaining experiments. Further study of consecutive components is a part of our future work.

\subsubsection{Healing monitoring:}

In this task, we evaluate changes of the metric $H$ in different stages of the tendon healing. Fig.~2 presents the time curves of the average $H$ metric for 9 patients, monitored in 10 timesteps distributed over the year of the tendon recovery, for all of the 10 MRI protocols. 

\begin{figure}[h!]
\centering
\includegraphics[width=0.8\textwidth]{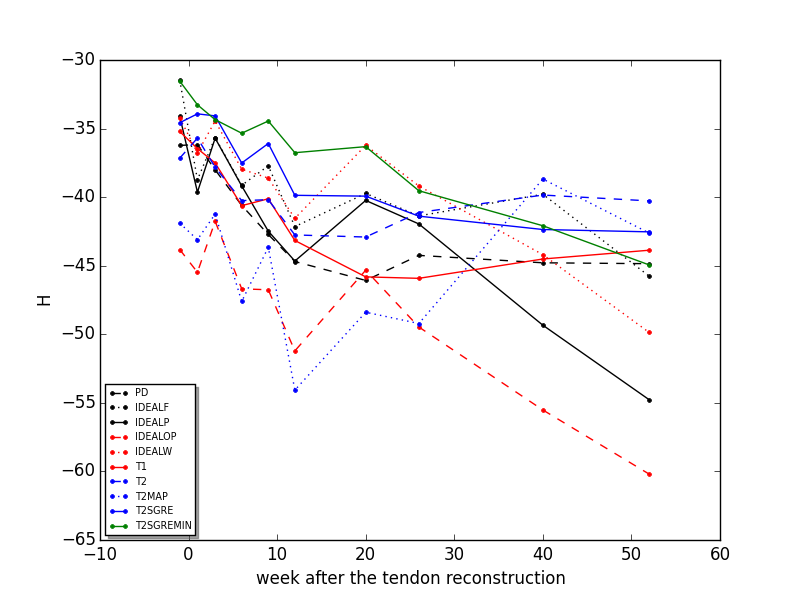}
\caption{Comparison of the curves of an average $H$ value over 9 patients monitored over the year of the tendon recovery for all of the 10 MRI protocols.}\label{fig:H}
\end{figure}

Except for the T2 mapping (T2MAP) with slices acquired from the raw signal data, all other protocols show a decreasing trend that can be interpreted as a negative difference between $H$ value before the reconstruction and at the end of the recovery period. According to the feedback provided by radiologists and medical professionals, all 9 curves correspond well with the real-life healing process assessment. In most cases the progress of healing is more pronounced at the beginning than in the following stages. However, one can also observe the fluctuations related to patient activity, diet and their obedience to the treatment prescription. Taking into account the above results, we select all 9 protocols in the following experiments. 

\subsubsection{Correlation of $H$ with the ground truth:}

Tab. 2 presents the values of Pearson correlation of the assessment done by the expert radiologist with our results obtained for 9 MRI protocols (excluding T2 mapping). The coefficients marked in bold are statistically significant with $p < 0.01, N=10$. For 4 protocols (i.e. PD, T1, T2, T2$^\ast$GRE) we can observe relatively high correlation with the ground truth in terms of the tendon and tissue edema size (TE and TisE), as well as the sharpness of the tendon edges (STE). Those parameters reach their maximum before the reconstruction and decrease with time. When they stabilize, the patients reach the end of the healing process. We also see that the parameter saturation coincides with plateauing of $H$ values for all 4 protocols. 

\begin{table}[t!]
\setlength{\tabcolsep}{2pt}
\centering
\caption{Pearson correlation between the ground truth parameters and the $H$ measure. Coefficients with significance p $<$ 0.01 are shown in bold. }
\label{correlation1}
\begin{tabular}{l||c|c|c|c|c|c|c|c|c}
     & PD                                   & IDELF & IDELP & IDELOP & IDELW & T1                                   & T2                                   & T2$^\ast$GRE                               & T2$^\ast$GREMIN \\ \hline \hline
SCT  & 0.69                                 & 0.42  & 0.53  & 0.59   & 0.55  & 0.66                                 & 0.60                                 & 0.74                                 & 0.4       \\ \hline
TT   & 0.52                                 & 0.71  & 0.46  & 0.30   & 0.37  & 0.55                                 & 0.40                                 & 0.38                                 & 0.45      \\ \hline
STE  & {\color[HTML]{333333} \textbf{0.87}} & 0.74  & 0.60  & 0.50   & 0.54  & {\color[HTML]{000000} \textbf{0.84}} & {\color[HTML]{000000} \textbf{0.81}} & 0.72                                 & 0.64      \\ \hline
TE   & {\color[HTML]{000000} \textbf{0.89}} & 0.53  & 0.63  & 0.62   & 0.58  & {\color[HTML]{000000} \textbf{0.81}} & {\color[HTML]{000000} \textbf{0.82}} & {\color[HTML]{000000} \textbf{0.82}} & 0.60      \\ \hline
TU   & 0.45                                 & 0.19  & 0.48  & 0.59   & 0.50  & 0.44                                 & 0.26                                 & 0.64                                 & 0.56      \\ \hline
TisE & {\color[HTML]{000000} \textbf{0.82}} & 0.51  & 0.61  & 0.65   & 0.58  & {\color[HTML]{000000} \textbf{0.84}} & 0.67                                 & {\color[HTML]{000000} \textbf{0.90}} & 0.73      \\ 
\end{tabular}
\end{table}

The TU parameter is assessed with the use of the sagittal slices. Thus, our neural network model is not able to successfully assess the healing phase using this parameter, as it is trained on axial cross-sections, not sagittal ones. We plan to further investigate the results of SCT and TT correlations and extend their analysis using more principal components.  
\subsubsection{Inter-protocol correlation:}
In this section, we analyse the correlation between results obtained with different MRI protocols to find a minimal subset of protocols that is needed by our method to provide satisfactory performance. We investigate 4 protocols that perform best in the previous task and present the inter-protocol correlations results in Tab.~3.

\begin{table}[t!]
\centering
\setlength{\tabcolsep}{12pt}
\caption{Inter-protocol correlation of PD, T1, T2 and T2$^\ast$ GRE MRI protocols. The results marked in bold have a significance level p $<$ 0.01.}
\label{my-label}
\begin{tabular}{l||c|c|c|c}
 & PD & T1 & T2 & T2$^\ast$GRE \\ \hline \hline
PD & 1.00 & {\color[HTML]{000000} \textbf{0.96}} & {\color[HTML]{000000} \textbf{0.90}} & {\color[HTML]{000000} \textbf{0.89}} \\ \hline
T1 & {\color[HTML]{000000} \textbf{0.96}} & 1.00 & {\color[HTML]{000000} \textbf{0.85}} & {\color[HTML]{000000} \textbf{0.92}} \\ \hline
T2 & {\color[HTML]{000000} \textbf{0.90}} & {\color[HTML]{000000} \textbf{0.85}} & 1.00 & 0.71 \\ \hline
T2$^\ast$GRE & {\color[HTML]{000000} \textbf{0.89}} & {\color[HTML]{000000} \textbf{0.92}} & 0.71 & 1.00  
\end{tabular}
\end{table}

\begin{figure}[t!]
\begin{center}
\includegraphics[width=0.93\textwidth]{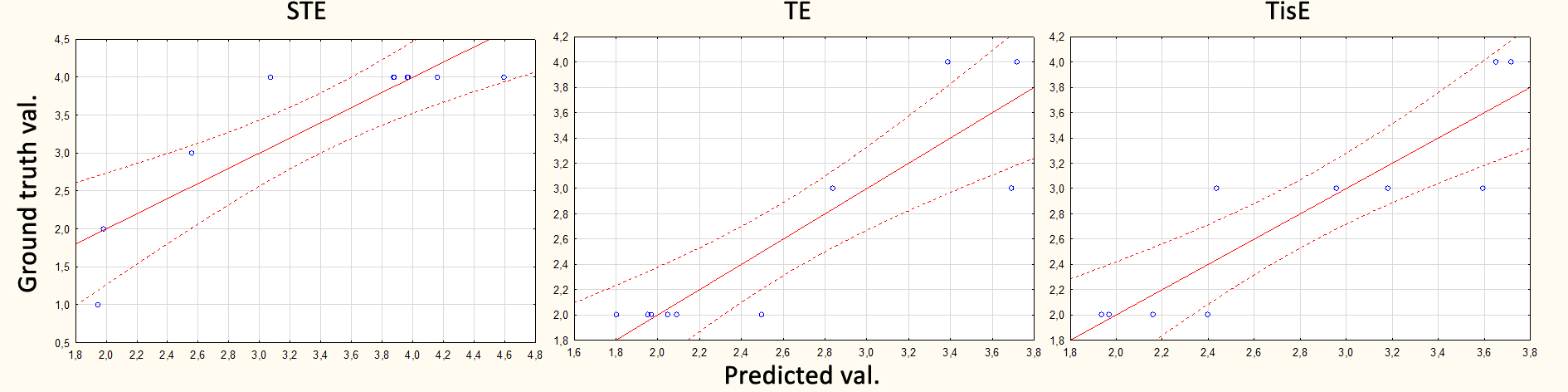}
\caption{Comparison of our model predictions with the ground truth. Our method predicts the correct outputs with error of less than a half of the score point.}\label{fig:regression}
\vspace{-0.6cm}
\end{center}
\end{figure}

The coefficients marked in bold are statistically significant with $p < 0.01, N=10$. Due to the fact that the PD and T1 protocols correlate strongly with each other, it is sufficient to select only one of them for our experiments. We identify two least correlated protocols, namely PD and T2$^\ast$GRE, and choose this pair. Although alternative methods for protocol selection exist, {\it e.g.} one can use backward feature elimination strategy, however, our initial experiments indicate that the performance of the proposed approach is sufficient.
\vspace{-0.3cm}
\subsubsection{Prediction of the healing parameters:}

Here, we test how accurate predictions of the ground truth we can obtain using a combination of the results for PD and T2$^\ast$GRE-based inference. We use a linear regression of the results from both protocols and analyze the predictions of STE, TE and TisE parameters. Fig.~3 presents the results of our study. 

The prediction error is below 1 score point in a 5-point scale for every case. This result confirms the validity of our method  and can justify using our method in the context of automatic assessment of the Achilles tendon healing. 

\section{Conclusions}
In this paper, we proposed to use convolutional neural networks to automatically assess the Achilles tendon healing phase. Currently, radiologists spend significant amount of time on analysing MRI images to manually evaluate tendond's condition. The results presented in this paper prove that methods based on deep neural networks can provide automatic, quantitative analysis of the 3D MRI scans. This, in turn, can lead to significant time savings and increase the efficiency of healing assessment without losing its precision. 

More precisely, we proposed a novel method for the tendon healing process assessment based on the pre-trained convolutional neural network. Our method allows to significantly improve the clinical workflow by minimizing the number of MRI protocols used to assess the healing and by providing objective, single value assessment for the current condition of the tendon. 

In this paper, we also presented the results confirming that our method allows to approximate 3 out of 6 assessment criteria with a human expert accuracy. At the same time, our method uses only 2 out of 10 typically used MRI protocols which can also prove the efficiency of the proposed method.

As future work, we plan to focus on approximating the remaining  parameters. Our idea is to use more principal components to compute the $H$ metric, as well as test different types of regression methods and extend our neural network to process also the data from sagittal cross-sections.

\subsection*{Acknowledgments}

{
The following work was part of {\it Novel Scaffold-based Tissue Engineering Approaches to Healing and Regeneration of Tendons and Ligaments (START)} project, co-funded by The National Centre for Research and Development (Poland) within STRATEGMED programme
(STRATEGMED1/233224/10/NCBR/2014).
}
\vspace{-0.3cm}

\bibliographystyle{ieeetr}
\bibliography{biblio}

\end{document}